\documentclass{bmcart}
\usepackage{graphicx}
\usepackage{setspace}
\usepackage{lineno}
\usepackage{soul}
\usepackage{xcolor}

\usepackage{hyperref} 

\usepackage[utf8]{inputenc} 
\usepackage{tabularx,booktabs}

\startlocaldefs
\endlocaldefs

\begin{document}

\begin{frontmatter}

\begin{fmbox}
\dochead{Research}


\title{Detecting DeFi Securities Violations from Token Smart Contract Code}


\author[
   addressref={aff1, aff2},                  
   corref={aff1},                       
   email={arianna.trozze@ucl.ac.uk}
]{\inits{A}\fnm{Arianna} \snm{Trozze}}
\author[
   addressref={aff2,aff3},
   email={\textbf{arianna.trozze@ucl.ac.uk}}
]{\inits{B}\fnm{Bennett} \snm{Kleinberg}}
\author[
    addressref={aff2},
    email={toby.davies@ucl.ac.uk}
]{\inits{T}\fnm{Toby} \snm{Davies}}

\address[id=aff1]{\orgname{Department of Computer Science, University College London}, 
\street{Gower Street}                  
\postcode{WC1E 6EA},                             
\city{London},                              
\cny{UK}                                    
}
\address[id=aff2]{\orgname{Department of Security and Crime Science, University College London},
\street{35 Tavistock Square},
\postcode{WC1H 9EZ}
\city{London},
\cny{UK}
}
\address[id=aff3]{\orgname{Department of Methodology \& Statistics, Tilburg University},
\street{Warandelaan 2},
\postcode{5037 AB}
\city{Tilburg},
\cny{Netherlands}
}




\end{fmbox}


\begin{abstractbox}
\begin{abstract} 
Decentralized Finance (DeFi) is a system of financial products and services built and delivered through smart contracts on various blockchains. In the past year, DeFi has gained popularity and market capitalization. However, it has also been connected to crime, in particular, various types of securities violations. The lack of Know Your Customer requirements in DeFi poses challenges to governments trying to mitigate potential offending in this space. This study aims to uncover whether this problem is suited to a machine learning approach, namely, whether we can identify DeFi projects potentially engaging in securities violations based on their tokens’ smart contract code. We adapt prior work on detecting specific types of securities violations across Ethereum, building classifiers based on features extracted from DeFi projects’ tokens’ smart contract code (specifically, opcode-based features). Our final model is a random forest model that achieves an 80\% F-1 score against a baseline of 50\%. Notably, we further explore the code-based features that are most important to our model’s performance in more detail, analyzing tokens’ Solidity code and conducting cosine similarity analyses. We find that one element of the code our opcode-based features may be capturing is the implementation of the SafeMath library, though this does not account for the entirety of our features. Another contribution of our study is a new data set, comprised of (a) a verified ground truth data set for tokens involved in securities violations and (b) a set of legitimate tokens from a reputable DeFi aggregator. This paper further discusses the potential use of a model like ours by prosecutors in enforcement efforts and connects it to the wider legal context.
\end{abstract}




\end{abstractbox}
%

\end{frontmatter}



\begin{doublespace}
\section*{Introduction}
Decentralized Finance (DeFi) refers to a suite of financial products and services delivered in a decentralized and permissionless manner through smart contracts\footnote{\footnotesize{Smart contracts are programs stored on a blockchain which automatically carry out specified actions when certain conditions are met \cite{RN539}.}} on a blockchain\footnote{\footnotesize{A blockchain is a secure, decentralized database comprised of entries called blocks, which are cryptographically connected to one another through a hash of the previous block, thereby ensuring its security and resistance to fraud. In the case of cryptocurrencies, blockchains serve as a decentralized, distributed public ledger that records all transactions \cite{RN533, RN540}. In this sense, blockchains underpin the ``decentralized'' nature of ``decentralized finance'', as they allow users to transact with one another in a trustless manner, without the need for an intermediary financial institution.}}, most commonly Ethereum. Its promoters have proclaimed it to be the future of finance \cite{RN499}, an assertion supported by the increase in its market capitalization by more than 8,000\% between May 2020 and May 2021 \cite{RN500}. Unfortunately, criminal activity in the DeFi ecosystem has also grown in parallel with its value. As of August 2021, 54\% of cryptocurrency fraud was DeFi-related, compared to only 3\% the previous year \cite{RN1535}. Furthermore, vast numbers of new DeFi projects are created every day and anyone is permitted to create one. Taken together, these present a challenge for law enforcement. The volume of projects, coupled with the magnitude of criminal offending, makes the development of an automated fraud detection method to guide investigative efforts particularly critical.
\par
Securities violations are one category of crime affecting the cryptocurrency space \cite{RN1505, RN1504, RN399}. Securities violations refer to offenses relating to the registration of securities and misrepresentations in connection with the purchase or sale of securities, including pyramid schemes and foreign exchange scams \cite{RN506, RN501}. Preliminary empirical research on decentralized exchanges (one of DeFi’s core product offerings) points to the prevalence of specific types of securities violations (such as exit scams\footnote{\footnotesize{Exit scams, also referred to as “rug pulls", involve developers of a project stealing all funds paid into or invested in their project \cite{RN537}.}}, advance fee fraud\footnote{\footnotesize{Advance fee fraud refers to a scammer convincing a victim to send them an amount of money in exchange for returning the original amount plus a premium. The fraudster simply takes the original funds \cite{RN538}.}}, and market manipulation) on these platforms \cite{RN480}, while others have chronicled securities violations like Ponzi schemes on decentralized applications (dApps) \cite{RN492}.\footnote{\footnotesize{DApps are the user interfaces of DeFi-based products and services.}} This limited empirical work suggests possible approaches for identifying general scam tokens and certain types of securities violations like Ponzi schemes in the wider cryptocurrency universe. While we acknowledge the realm of work using opcode-based features to identify malicious activity (see, for example, \cite{santos_opcode_2013}), research has not yet explored automated detection a) of securities violations overall (rather than specific types of securities violations like scam tokens or Ponzi schemes, b) across a broader subspace of the DeFi ecosystem (i.e., ERC-20 tokens on all DeFi platforms, instead of a single decentralized exchange); or c) alongside detailed analyses of the ERC-20 tokens' smart contract code. We consider an automated approach to be preferable because of the sheer volume of DeFi projects that exist and are being created. 
\par 
Against this background, we seek to answer the following research questions: (1) is a machine learning approach appropriate for identifying DeFi projects likely to be engaging in violations of U.S. securities laws?\footnote{\footnotesize{Developers write smart contracts in a high-level programming language called Solidity \cite{cai_decentralized_2018}. Smart contracts are responsible for DeFi’s application infrastructure, as well as for creating cryptocurrency tokens themselves.}} and (2) what are the reasons, at feature level, such a model is or is not successful for this classification problem? This study presents and critically evaluates the first method for automated detection of various types of securities violations in the DeFi ecosystem on the basis of their token’s smart contract code, providing a tool which may identify starting points for further investigation. The contributions of this study are as follows:

\begin{itemize}
  \item We build a classifier to detect DeFi projects committing various types of securities violations. Our work is the first to expand existing machine learning-based classification models to encompass multiple types of securities violations.
  \item We use and make available a new data set of violations verified by court actions.
  \item Our work is the first to prioritize the explainability of classification decisions in terms of opcode-based features.
\end{itemize}

Finally, the results of this study contribute to the theory and practice of financial markets. Forecasting, detecting, and deterring financial fraud is critical to maintaining overall financial stability \cite{shams_detection_2021}. In particular, ``frauds harm the integrity of financial markets and disrupt the mechanism of efficient allocation of financial resources'' \cite{shams_detection_2021}. This is particularly pertinent in the cryptocurrency space \cite{shams_detection_2021}, especially as these markets become more entwined with traditional ones \cite{wang_short-_2022}.

\section*{Decentralized Finance (DeFi)}
DeFi refers to a collection of financial products and services made possible by smart contracts built on various blockchains, most commonly the Ethereum blockchain. DeFi offers traditional financial products and services, such as loans, derivatives, and currency exchange, in a decentralized manner through smart contracts. DeFi is an open source, permissionless system, that is not operated by a central authority. Rather than transacting with one another through an intermediary like a centralized exchange, users' interactions occur through dApps, created by smart contracts, on a blockchain \cite{schar_decentralized_2021}. In this section we describe our DeFi system model for this research and briefly outline its main components. As it is the subject of our research, we focus our explanation on the Ethereum-based DeFi space, though DeFi, of course, exists on various blockchains.
\par
\subsection*{Ethereum-based DeFi System Model}
\begin{figure}[h!]
    \begin{center}
        \includegraphics[width=7cm, keepaspectratio]{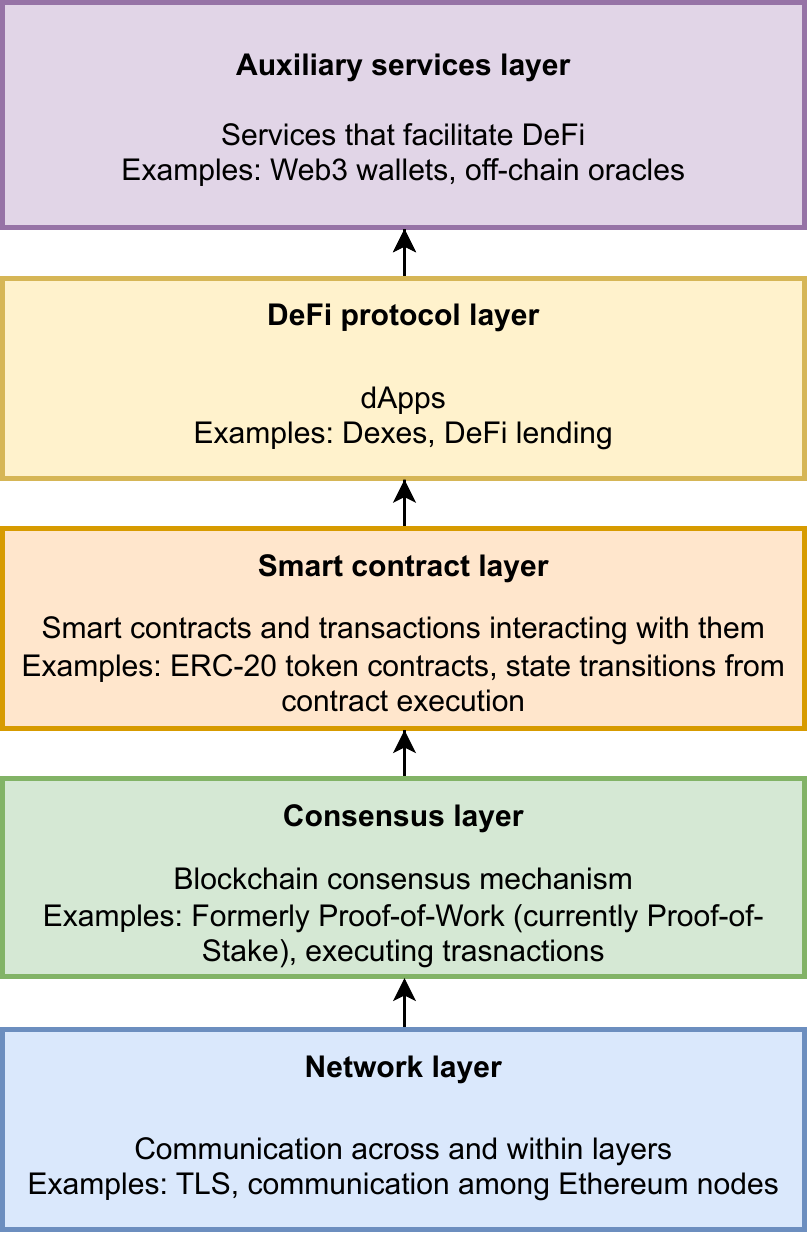}
        \begin{singlespace}
        \caption{\csentence{Ethereum-based DeFi System Model.} Five layers of the Ethereum-based DeFi system (from bottom to top): network layer, blockchain consensus layer, smart contract layer, DeFi protocol layer, and auxiliary services layer.}
        \end{singlespace}
  \end{center}
  \label{fig1}
\end{figure}

Before we explain DeFi in more detail, we define our system model. The Ethereum-based DeFi system model can be conceptualized as a five-layer system, consisting of a network layer, a blockchain consensus layer, a smart contract layer, the DeFi protocol layer, and an auxiliary services layer \cite{zhou_sok_2022}. The network layer is concerned with communicating data across and within the various layers. It involves several elements, including network communication protocols and the Ethereum network; in particular, it includes communication among Ethereum peers/nodes. The consensus layer refers to the Ethereum blockchain's consensus mechanism \cite{zhou_sok_2022}. At the time of our research, this was still Proof-of-Work \cite{RN502}. The consensus layer also encompasses nodes' actions which rely on the consensus mechanism such as ``data propagation'', ``data verification'', executing transactions, and mining blocks. While the first two layers are implied in our research, this study primarily focuses on the smart contract layer. This includes the smart contract code that creates the ERC-20 tokens from which we derive our data set, and that creates the dApps that use them. The smart contract layer also includes transactions executed by smart contracts, the Ethereum Virtual Machine (EVM) state, and the state transition upon execution of DeFi transactions. The DeFi protocol layer refers to decentralized applications with which users interact, and the auxiliary service layer involves services that facilitate DeFi's functioning, such as wallets and off-chain oracles. We describe these elements in more detail below and provide a visual representation thereof in Figure \ref{fig1}.

\subsection*{Ethereum}
Ethereum functions as a distributed virtual machine and is the platform on which much of the DeFi ecosystem currently runs. This paper focuses on explaining the features of Ethereum which are most relevant to DeFi’s functioning.\footnote{\footnotesize{For further details on Ethereum, see the Ethereum Yellow Paper \cite{RN502}.}}
\par
In addition to holding balances, Ethereum accounts can store smart contract code and other information. A smart contract is a computer program that automatically carries out certain actions when specified conditions—such as payments—are met \cite{RN540}. The smart contract code is immutable and publicly available on the blockchain. Smart contracts allow parties who do not trust one another to enter into contracts—rather than trusting each other or a third party to execute the contract, smart contracts ensure the terms will be executed as coded into the contract \cite{RN539}.\par
Before smart contracts are executed, they must be compiled into bytecode to be deployed and understood by the Ethereum Virtual Machine (EVM). Once compiled, Ethereum smart contracts take the form of a string of numbers (bytecode). The EVM is a stack-based environment\footnote{\footnotesize{In stack-based programming, ``all functions receive arguments from a numerical stack and return their result by pushing it on the stack.'' These specific functions come from a set of pre-defined functions \cite{RN534}.}}, with a 256-bit stack size. It reads the bytecode as operational codes (opcodes) which are, essentially, a set of instructions (from a set of 144 possible instructions). The opcodes include actions like retrieving the address of the individual interacting with the contract, various mathematical operations, and storing information \cite{RN503, RN502}. 
\par
\subsection*{Tokens}
Many DeFi projects also have associated tokens created by smart contracts, which either entitle holders to something within the dApp (analogous to a video game’s in-game currency) or serve as ``governance tokens''. For example, UNI is the Uniswap decentralized exchange’s governance token \cite{RN515}. Another example is the SCRT token which, in addition to being a governance token, is required to pay transaction fees on the Secret network \cite{RN516}. Holders of governance tokens can vote on the future of projects; their voting power is proportional to the amount of governance tokens they hold. Most non-NFT DeFi tokens on Ethereum follow the ERC-20 (Ethereum Request for Comment) standard, which facilitates interoperability among projects. The ERC-20 standard allows for token capabilities like transferring among accounts, maintaining balances, and the supply of tokens, among others \cite{RN504}. In this paper, we focus on ERC-20 tokens.
\par 
\subsection*{dApps}
Developers create dApps which serve as the interfaces to execute these smart contracts. DeFi’s current core product offerings—including decentralized exchanges (dexes), lending products, prediction markets, insurance, and other financial products and services—are delivered through dApps \cite{RN438}. Table \ref{tab1} describes the primary products that make up the DeFi ecosystem. 
\par
\begin{figure}[!htpb]
    \begin{center}
	\includegraphics[height=17cm, keepaspectratio]{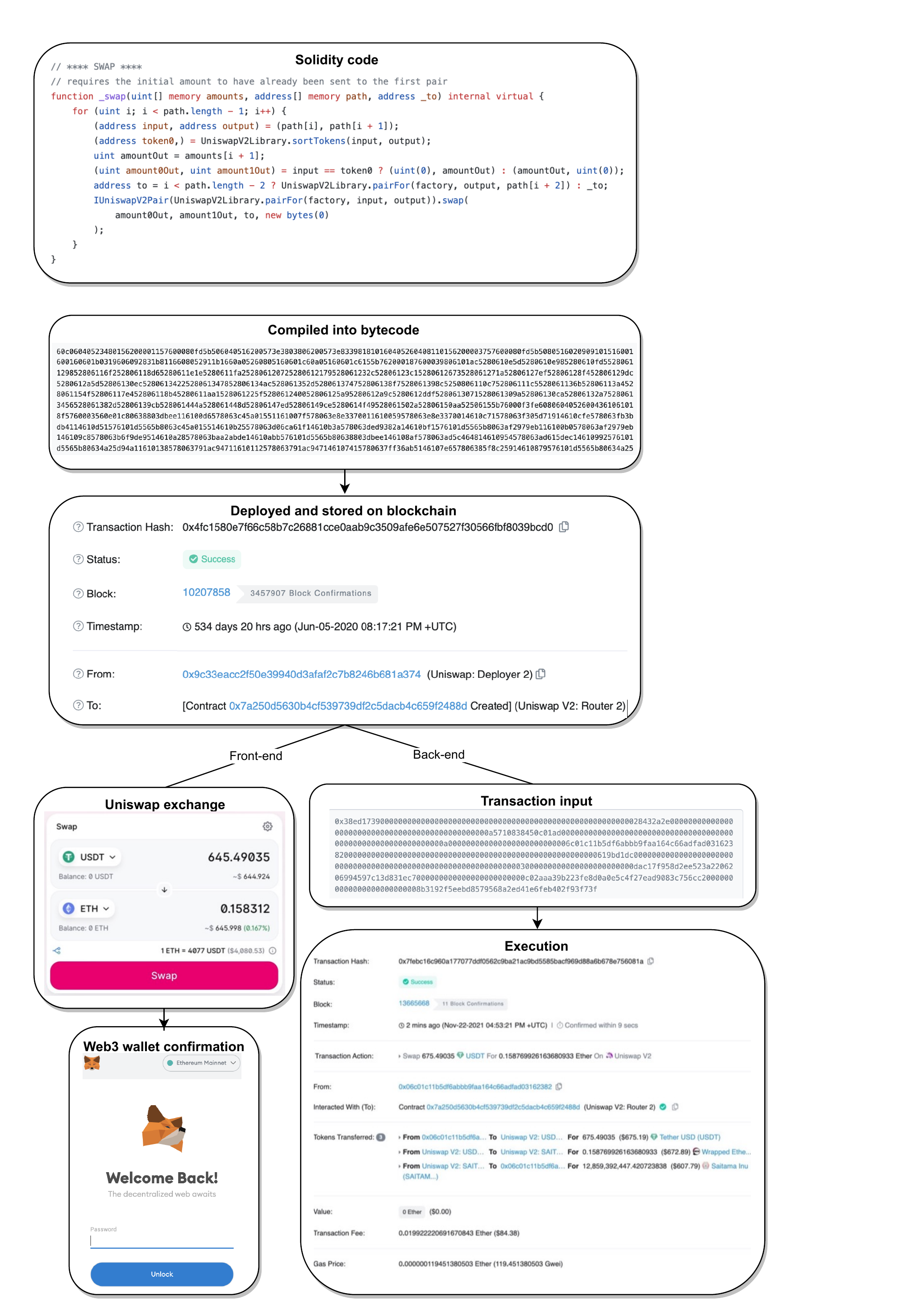}
        \begin{singlespace}
            \caption{\csentence{DApp creation and functioning.} The first box contains an excerpt from the Solidity code for the popular dex Uniswap's exchange function. The second box shows the same code compiled into EVM-readable bytecode. The third box shows the transaction which deployed this code to the Ethereum blockchain, thereby creating the dApp. The branch of Figure \ref{fig2} labelled ``front-end'' shows Uniswap's user interface for a sample exchange operation of 645.49035 USDT to ETH. Below the exchange interface is a screenshot of the Web3 wallet MetaMask. To execute a transaction like the exchange depicted above, the user must connect their Web3 wallet to the relevant dApp via the wallet's browser extension. From there, they can approve the transaction. After the transaction is executed, the user's MetaMask wallet automatically reflects the new balances of these cryptocurrencies. On the back-end, the aforementioned exchange takes the form of bytecode (depicted on the ``back-end'' branch of the figure), which is executed by the EVM. the final box shows the hash of the executed transaction exchanging USDT for ETH. Full details of this exchange can be found at \href{https://etherscan.io/tx/0x7febc16c960a177077ddf0562c9ba21ac9bd5585bacf969d88a6b678e756081a}{https://etherscan.io/tx/0x7febc16c960a177077ddf0562c9ba21ac9bd5585bacf969d88a6b678e756081a}. The full Solidity source code for the Uniswap V2 smart contract can be found at \href{https://etherscan.io/address/0x7a250d5630b4cf539739df2c5dacb4c659f2488d\#code}{https://etherscan.io/address/0x7a250d5630b4cf539739df2c5dacb4c659f2488d\#code}.}
        \end{singlespace}
    \end{center}
  \label{fig2}
  \end{figure}

Figure \ref{fig2} depicts the process of dApp creation and execution through smart contracts, using Uniswap (a popular dex) as an example. As depicted in Figure \ref{fig2}, DeFi users must have a Web3 software wallet to hold DeFi tokens and interact with dApps. These wallets can be thought of as akin to a mobile banking application and exhibit similar features (sending transactions, showing balances, etc.). The difference is that, unlike with a banking application, users retain custody of their funds and can send transactions and execute other functions directly, rather than through an intermediary institution \cite{RN1538}. Using cryptographic digital signatures, users approve connections to their Web3 wallets, ``sign in'' to dApps, and approve interactions with the smart contracts on these platforms through their wallet.

\begin{singlespace}
\begin{table}[htpb]
\caption{Overview of DeFi products.}
    \begin{tabularx}{\textwidth}{lX}\hline
        \label{tab1}
        \textbf{Decentralized exchanges} & These services allow users to exchange cryptocurrencies using liquidity provided by other users, generally through Automated Market Makers, which algorithmically set prices \cite{xu_sok_2022}. Participants can provide liquidity to liquidity pools for certain pairs of cryptocurrencies and receive a Liquidity Provider token for doing so. They can ``stake'' this token (i.e., lock it into the system and agree not to withdraw it for a certain period) and earn interest on it, usually paid in the decentralized exchange’s governance token (referred to as ``yield farming''). The return on investment for these yield farms may be in the hundreds or even thousands of percent. Participants can stake governance tokens in ``pools'' and earn further rewards. This incentivizes users to provide liquidity to keep the exchanges running.\footnote{\footnotesize{For a more detailed discussion of decentralized exchanges, see \cite{RN480}.}} \\ \hline
        \textbf{DeFi lending} & Loans are issued through smart contracts rather than intermediaries and use cryptocurrencies as collateral \cite{bartoletti_sok_2020}. Loans are often issued in stablecoins—cryptoassets whose value is pegged to government-issued fiat currencies—and interest rates tend to be set algorithmically \cite{RN1516}. Users can earn interest for providing liquidity for loans and earn fees from loans. One primary DeFi lending innovation is flash loans, which are loans issued and repaid in a single transaction. Because they are issued and repaid in one transaction, they do not require collateral \cite{RN537}.\\ \hline
        \textbf{Prediction markets} & These allow users to bet on real-world outcomes—such as sporting events or elections—through smart contracts \cite{RN1520}. Prediction markets rely on blockchain oracles, which are external sources of information that determine the outcome of the prediction market \cite{RN537}. Based on this information about the outcome, the smart contract releases the appropriate funds to the winners  \cite{RN1520}. \\ \hline
        \textbf{DeFi insurance} & DeFi insurance community members serve as underwriters and share in premiums paid to the protocol. Holders of the project’s governance token vote on claims payouts. DeFi insurance remains a nascent industry, but some companies are attempting to handle claims directly through smart contracts. So far, DeFi insurance tends to insure only other DeFi protocols  \cite{RN1524}. \\ \hline
        \textbf{Other financial products} & A range of other financial products, including those not usually available to retail investors, can be implemented within DeFi. These include derivatives trading, margin trading, and other securities \cite{RN1537}.   \\\hline
    \end{tabularx}
\end{table}
\end{singlespace}

\doublespacing
\subsection*{U.S. Securities Laws}

An understanding of U.S. securities law is necessary prior to defining our DeFi threat model. DeFi has raised alarm in regulatory circles due to concerns over DeFi tokens’ potential conflict with existing U.S. securities laws \cite{RN507}. U.S. securities are primarily governed at the federal level by the Securities Act and the Exchange Act, though the Sarbanes-Oxley Act, Trust Indenture Act, Investment Advisors Act, and Investment Company Act are also relevant. Enforcement of these laws is the responsibility of the Securities and Exchange Commission (SEC) and the Financial Industry Regulatory Authority (FINRA) \cite{RN501}.
\par
The Securities Act relates to the offer and sale of securities. One of the key provisions which has been charged in cryptocurrency cases is Section 5, which requires the registration of the offer and sale of securities and stipulates specific provisions thereof \cite{RN501}.\footnote{\footnotesize{For cryptocurrency case law involving the Securities Act, see \cite{RN517, RN518, RN521}.}} Other sections detail the required registration information\footnote{\footnotesize{Sections 7 and 10.}} and exemptions\footnote{\footnotesize{Section 3, Section 4, Regulation S, Rule 144A, Regulation D, Rule 144, Rule 701, Section 28.}} \cite{RN501}. Various SEC enforcement actions have successfully argued that ERC-20 tokens are securities (see, for example, \cite{SEC_lbry}), arguing that they constitute investment contracts \cite{us_securities_and_exchange_commission_framework_2019}. For further details on the application of the Howey Test (the SEC’s criteria for determining whether something is a security), as it relates to digital assets, see \cite{us_securities_and_exchange_commission_framework_2019}.
\par
The Exchange Act specifies public companies’ reporting requirements and regulates securities trading through securities exchanges. It also handles securities fraud; under section 10(b) and 10b-5, fraud and manipulation in relation to buying or selling securities is illegal. One cannot make false or misleading statements (including by omission) in relation to the sale or purchase of securities, including those exempt from registration under the Securities Act. The registration of securities, securities exchanges, brokers, dealers, and analysts is covered in Sections 12 and 15 of the Exchange Act \cite{RN501}. Section 12 regulates registration of initial public offerings, which is relevant for cryptocurrency initial coin offerings (ICOs).\footnote{\footnotesize{For cryptocurrency case law involving the Exchange Act, see \cite{RN517, RN518, RN521}.}} Finally, Section 13 relates to companies’ reporting obligations under the Exchange Act \cite{RN501}.
\par
In practice, in addition to various registration and reporting violations, securities laws in the U.S. tend to cover the following fraudulent conduct: high yield investment programs, Ponzi schemes, pyramid schemes, advance fee fraud, foreign exchange scams, and embezzlement by brokers, among others \cite{RN506}.\footnote{\footnotesize{For definitions of these offenses, see \cite{RN538}}}. Financial frauds have been shown to impact financial stability, market integrity, and resource allocation and, in the case of cryptocurrency frauds, to have an impact on markets in traditional finance \cite{shams_detection_2021, xin_economic_2018}.

\subsection*{Threat Model}
We define our threat model in line with the U.S. securities laws described above. With that in mind, an ``incident'' is any activity that is in violation of these laws, such as failing to register a token as a security or an ICO or committing securities fraud. These actions may be the result of intentionally malicious behavior or naivete. Both cases ``result in an unexpected financial loss'' to users \cite{zhou_sok_2022}. While we do not have an estimate of all of the losses individuals incurred as a result of DeFi securities violations, we know, for example that the Finiko ponzi scheme took \$1.1 billion from victims in 2021 and rug pulls stole \$2.8 billion worth of funds from victims in 2021 \cite{chainalysis_2022_2022}. These incidents occur at the smart contract layer, the protocol layer, and the auxiliary layer of the DeFi system. In terms of the smart contract layer, this includes both the creation of the ERC-20 token itself (in the case of registration violations), as well as any malicious elements coded into the smart contracts such as Ponzi schemes, advance feed fraud, or certain types of exit scams. At the protocol level, the primary attack vector would be market manipulation \cite{zhou_sok_2022}. Finally, at the auxiliary layer, we see both ``operational vulnerability'' (such as price oracle manipulation) and ``information asymmetry'' (such as smart contract honeypots) \cite{zhou_sok_2022}. Information asymmetry is primarily at play in the case of securities fraud. Users are often unable to (or do not take the time to) analyze DeFi protocol smart contracts (and related security risks) prior to allowing them to utilize their assets. Users' ``understanding of a contract operation'' is more likely to come from projects' marketing materials than the contract source code itself \cite{zhou_sok_2022}. 

The above threat model involves various assumptions. First, we assume that, based on the classification of the tokens used to construct our data set by the SEC, the DeFi tokens in question are securities under U.S. law. As discussed above, precedent has been established in this regard. Notably, this also means that many otherwise legitimate projects may be operating contrary to U.S. securities laws by virtue of the fact that they are not appropriately registered. The second assumption we make is that, in the case of frauds coded into the smart contract code, the developers of the DeFi tokens violating securities laws are behaving maliciously, rather than their violations being the result of errors. For this reason, patching or fixing smart contract code (as discussed in
\cite{ 263790}
and \cite{ferreira_torres_elysium_2022}) would not be a suitable way to allay this threat. Registration violations may result from either malicious intent or naivete.

\begin{figure}[h!]
    \begin{center}
        \includegraphics[width=12.25cm]{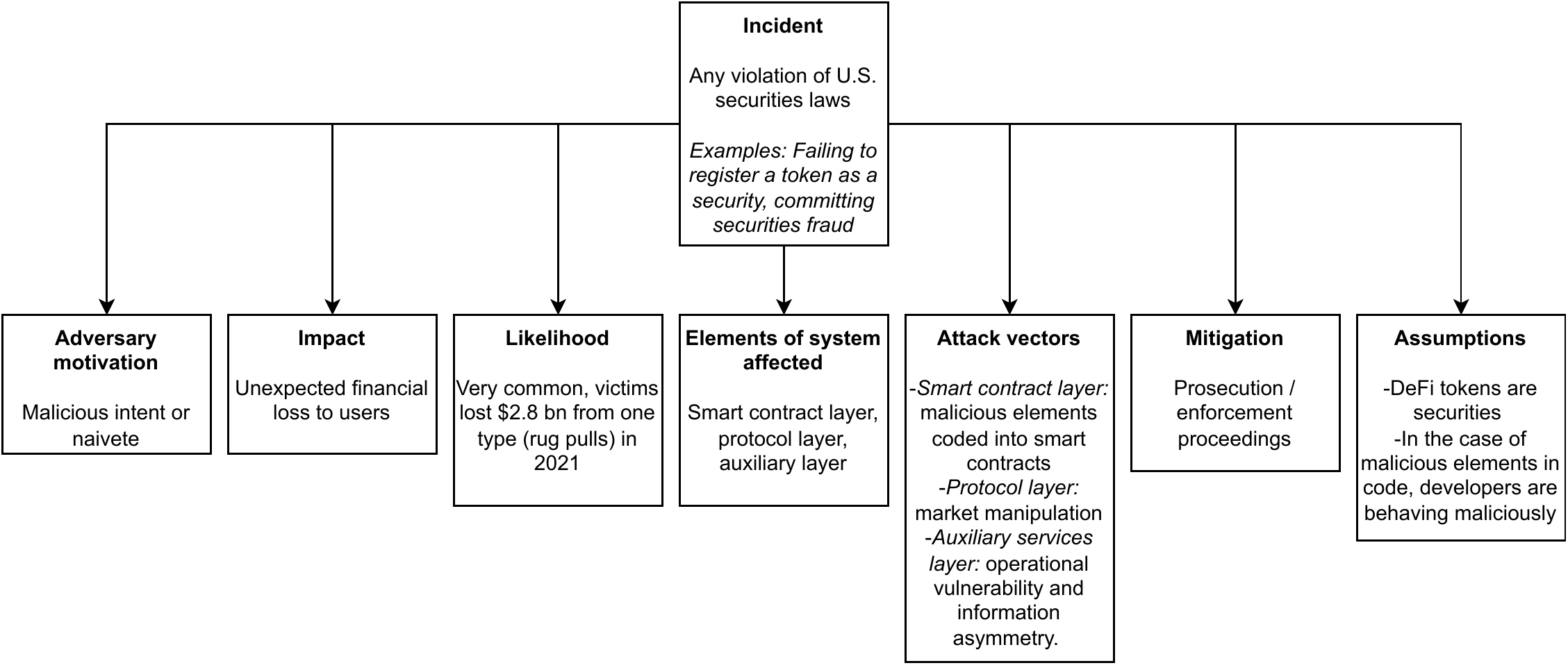}
        \begin{singlespace}
            \caption{\csentence{Threat model.} Threat model for violations of U.S. securities laws including attacker motivation, incident impact, elements of the system affected, attack vectors, mitigation, and assumptions.}
        \end{singlespace}
    \end{center}
  \label{fig3}
\end{figure}

Finally, we assume that prevention measures have failed in these instances and, therefore, the primary course of justice is detection and prosecution. While prevention is certainly preferable, it is not possible to prevent all crimes and, therefore, detection and prosecution remain an important way to remedy the threats discussed above. Figure \ref{fig3} offers a visual representation of our threat model.

\section*{Related Work on Detecting Fraud on Ethereum}
Previous studies have used machine learning to detect certain specific types of securities violations and fraud on Ethereum. The most common type of securities violation examined in the literature is smart contract Ponzi schemes \cite{RN486, RN481, RN1541, RN491, RN487, RN492, RN395, RN484, RN490, RN489}. The various machine learning algorithms these studies employ to identify such Ponzi schemes—and their relative performance—can be found in Table \ref{tab2}. We acknowledge the existence of various sequential machine learning studies in other contexts (many of which feature more sophisticated classification models). However, in this article, we limit our review to studies which apply sequential machine learning techniques to detect fraud on Ethereum, as the classification problems they seek to solve are most similar to ours.
\par

\begin{table}[htpb]
\caption{Related work detecting Ethereum Ponzi schemes.}
\begin{tabularx}{\textwidth}{c p{3cm} p{2cm} p{5cm}}\hline
\label{tab2}
\textbf{Study} & \textbf{Method} & \textbf{Features } & \textbf{Performance} \\ \hline
 \cite{RN486} & Semantically-aware classifier that includes ``a heuristic-guided symbolic execution technique'' & Code-based & Precision: 100\% \newline Recall: 100\% \newline F1: 100\% \\ \hline
\cite{RN491} & ``Anti-leakage'' model based on ordered boosting & Code-based & Precision: 95\% \newline  Recall: 96\% \newline F1: 96\% \\ \hline
\cite{RN487} & Deep learning model & Code-based & Precision: 96.3\%\newline Recall: 97.8\% \newline F1: 97.1\% \\ \hline
\cite{RN492} & Long-term short-term memory neural network & Transaction-based & Precision: Between 88.2\% and 96.9\% for different types of contracts \newline Recall: Between 81.6\% and 97.7\% for different types of contracts\newline F1: Between 85\% and 96.7\% for different types of contracts \\ \hline
\cite{RN490} & Long-term short-term memory neural network & Code- and transaction-based & Precision: 97\% \newline Recall: 96\% \newline F1: 96\% \\ \hline
\cite{RN484} & Heterogeneous Graph Transformer Networks & Code- and transaction-based & F1: Between 78\% and 82\% for fraudulent smart contracts and 87\% and 89\% for normal smart contracts for different classification tasks \\ \hline
\cite{RN489} & LightGBM & Code- and transaction-based & Precision: 96.7\% \newline Recall: 96.7\% \newline F1: 96.7\% \\ \hline
\cite{RN1541} & XGBoost & Code- and transaction-based & Precision: 94\% \newline Recall: 81\% \newline F1: 86\% \\ \hline
\cite{RN395} & Decision trees, random forest, stochastic gradient descent & Code- and transaction-based & Precision: Between 90\% and 98\% for different models \newline Recall: Between 80\% and 96\% for different models\newline F1: Between 84\% and 96\% for different models \\ \hline
\cite{RN1540}& Random forest & Code- and transaction-based & Precision: Between 64\% and 95\% for different features \newline Recall: Between 20\% and 73\% for different features\newline F1: Between 30\% and 82\% for different features \\ \hline
\end{tabularx}
\end{table}

Previous work on smart contract Ponzi schemes has examined code-based features \cite{RN486, RN487, RN491}; transaction-based features \cite{RN492}; or both \cite{RN490, RN484, RN395, RN489, RN1541, RN1540}. Code-based features include the frequency with which each opcode appears in a smart contract and the length of the smart contract bytecode \cite{RN395}. Transaction- and account-based features refer to the number of unique addresses interacting with the smart contract, and the volume of funds transferred into and out of the smart contract, among others \cite{RN395}. Notably, one study \cite{RN486} identifies four specific Ponzi scheme typologies based on bytecode sequences.
\par
In addition to smart contract Ponzi schemes, other studies examining smart contracts use machine learning to detect general fraud and scams 
\cite{RN485, RN496, RN398, RN494, RN480, fan_smart_2022}; advance fee fraud \cite{RN1542}; smart contract honeypots \cite{RN487, RN397}; ICO scams \cite{RN493, RN435}; and ``abnormal contracts'' causing financial losses 
\cite{aljofey_feature-based_2022}
. Notably, one study
\cite{aljofey_feature-based_2022}
also uses contract source code-based features, as well as opcode and transaction-based ones.
\par
Most existing work in this field examines Ethereum smart contracts in general, but some specifically refer to dApps and DeFi in their work \cite{RN491, RN492, RN490, RN494, RN480}, though they do so to varying degrees and, at times, conflate DeFi with Ethereum more broadly. Notably, one study \cite{RN492} uses machine learning to classify different types of dApp smart contracts into their various categories, including those for gaming, gambling, and finance, among others. 
\subsection*{Gaps and Issues}
While the results of the studies described in Table 2 suggest that approaches of this nature can perform well for this task, a number of points of caution have also been raised. Literature concerning smart contract Ponzi scheme detection points to issues of overfitting, due to the imbalance of classifications in many data sets \cite{RN491}. Studies have addressed this using the over- and under-sampling techniques \cite{RN486, RN491, RN490, RN489}. Other scholars \cite{RN486} criticize the interpretability of results based on opcode features, i.e., why the presence of certain opcodes would point to criminality. Perhaps most fundamentally, however, these studies have given little consideration to whether machine learning techniques are actually necessary for this task; or, at least, superior to other, possibly simpler, approaches. While the methods applied undoubtedly show high performance, it remains possible that similar metrics could be achieved without recourse to these kinds of techniques.
\par
Another issue with previous work is the repeated use of two particular data sets \cite{RN1541, RN508}. Of the studies cited above, four use the Bartoletti et al. \cite{RN508} data set  \cite{RN486, RN487, RN484, RN395} and two use the Chen et al. \cite{RN1541} one \cite{RN490, RN489}, while one study combined both data sets and added additional data \cite{RN491}. While using the same data sets may be useful for comparing performance, it may be less useful in practice for combating fraud, with any shortcomings of these data sets having a polluting effect on the literature. Indeed, when manually inspecting the Bartoletti et al. \cite{RN508} data set, one study \cite{RN486} identified issues involving duplication and bias. Finally, though used to classify general fraud rather than Ponzi schemes, one article uses proprietary company data \cite{RN494}, which hinders reproducibility and evaluations of results. 
\par
The majority of existing related work uses smart contracts in general as opposed to ERC-20 token smart contracts (as \cite{RN480} and we do). The only other work that specifically examines DeFi token smart contracts using machine learning is \cite{RN480}. However, their work focuses on scam tokens in general (rather than securities violations) and on a single dApp (the dex Uniswap).
\par
\subsection*{Aims of this Paper}
This paper seeks to fill these gaps by a) evaluating whether a machine learning approach is appropriate for identifying DeFi projects likely to be engaging in securities violations; b) examining  securities violations more comprehensively (rather than just scam tokens or Ponzi schemes); c) investigating these violations across Ethereum-based DeFi, rather than specific sub-spaces, like decentralized exchanges; and d) examining the code-based features identified by our model in order to better explain its performance. We also develop an entirely new data set of violating and legitimate tokens.

\section*{Method}

\begin{figure}[!htpb]
    \begin{center}
        \includegraphics[height=21cm]{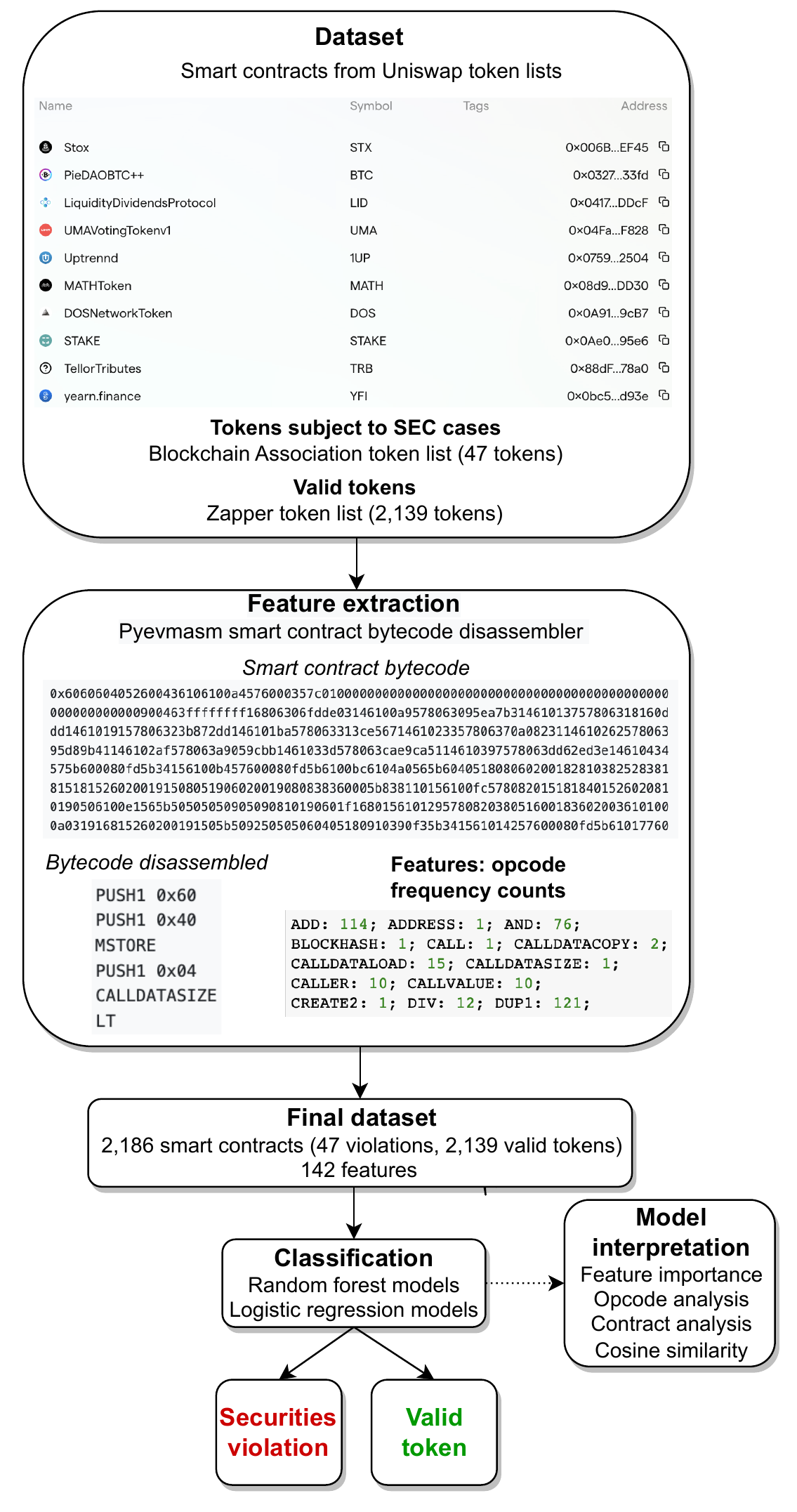}
        \begin{singlespace}
            \caption{\csentence{Methods for detecting securities violations from DeFi token smart contract code.}}
        \end{singlespace}
    \end{center}
  \label{fig4}
\end{figure}

This paper adapts its methods from prior research on the detection of Ethereum smart contract Ponzi schemes, adapting an approach that performed well in that context \cite{RN395} and applying it to DeFi projects engaging in securities violations. We build various classification models based on features extracted from DeFi tokens’ smart contract code to classify the tokens into two categories: securities violations and legitimate tokens. Figure \ref{fig4} provides a summary of our methods.

\subsection*{Data Collection}

To answer our question of whether it can be determined if a project may be engaging in securities violations from its token’s smart contract code using machine learning, we first required ground truth sets of both securities violations and legitimate tokens. One source of such information is provided by token lists compiled by DeFi projects or companies around particular themes. One function of token lists is to help combat token impersonation and scams; reputable token lists provide users with some assurance that the tokens that appear are not fraudulent. Uniswap posts lists which are contributed by projects in the community; users generally follow lists from projects they trust \cite{RN509}. The lists contain information like project websites (important for avoiding phishing attempts), symbols, and their smart contract addresses.

\subsubsection*{Securities Violations}

The Blockchain Association (BA), a lawyer-led blockchain lobbying organization, has created one such list of ERC-20 tokens which have been subject to U.S. SEC enforcement actions.\footnote{\footnotesize{\href{https://tokenlists.org/token-list?url=https://raw.githubusercontent.com/The-Blockchain-Association/sec-notice-list/master/ba-sec-list.json}{https://tokenlists.org/token-list?url=https://raw.githubusercontent.com/The-Blockchain-Association/sec-notice-list/master/ba-sec-list.json}}} At the time of our study, this list contained 47 tokens and these represent our ground truth for projects engaging in securities violations. Many of these actions involve Initial Coin Offerings (ICOs), primarily in the context of companies or individuals failing to register their token as a security when it was required and/or making fraudulent misrepresentations in connection with said token (for example, Coinseed Token, Tierion, ShipChain SHIP, SALT, UnikoinGold, Boon Tech, and others)  \cite{us_securities_and_exchange_commission_cyber_2022}. Other violations include Ponzi schemes (RGL, for example) and market manipulation (Veritaseum, for example)  \cite{noauthor_securities_2019-1, noauthor_securities_2019}. The defendants in these cases are distinct, thereby supporting the independence of the tokens in our violations data set. We acknowledge the limitations of using such a small set, but this comprises all SEC actions involving DeFi tokens to date. Therefore, it makes for a more credible ground truth data set than attempting to find individual investment scams and securities violations on blockchain forums (as other data sets do, including \cite{RN508}) in that it is more systematic and does not involve any subjective judgment of wrongdoing. It is worth noting that the nature of the list itself highlights the need for a systematic detection method: most of the actions either derived from the U.S. government whistleblower program, or were well-publicized scams, suggesting that enforcement is currently reliant on these sources. \cite{us_securities_and_exchange_commission_2021_2021}.

\subsubsection*{Legitimate Projects}

The nature of the DeFi industry means that a substantial proportion of tokens in general may be of questionable validity—even if they have not been formally identified as violations—which poses a challenge when building a data set of legitimate tokens. Were we to simply take a random sample of all projects, for example, it is likely that this would contain problematic tokens, compromising our analysis. For this reason, we took an alternative approach, including only tokens for which we had some evidence of credibility. To do this we use the token list maintained by the DeFi platform Zapper.\footnote{\footnotesize{\href{https://tokenlists.org/token-list?url=https://zapper.fi/api/token-list}{https://tokenlists.org/token-list?url=https://zapper.fi/api/token-list}}} Zapper serves as a DeFi project aggregator which allows users to monitor their liquidity provision, staking, yield farming, and assets across different DeFi protocols. As of November 2021, Zapper had over a million total users, \$11 billion worth of transaction volume, and raised \$15 million in venture funding \cite{RN510}. While they do not claim to provide financial advice to users, they make an effort to internally vet projects they list, opting for those with audited contracts and reputable teams \cite{RN511}. To be clear, inclusion in this list does not provide any indication of the ``quality'' of a token—it is not analogous to a list of ``blue chip'' stocks—but simply an indication of authenticity. The Zapper list contains 2,146 ERC-20 tokens, and we used this as our ground truth of legitimate tokens. We acknowledge that this may not be as representative of DeFi tokens in general as a random sample, but this is the best available source of tokens which have some marker of credibility. 

\subsubsection*{Final Data set}

We extracted the smart contract addresses for the ERC-20 tokens on both of these lists and combined them into a single data set, with a binary indicator added to flag violations. This gave us an initial data set of 2,193 smart contract addresses. Seven tokens were present on both lists. These likely represent tokens which are otherwise legitimate, but violated U.S. securities laws by failing to register as securities. Since the Blockchain Association list is verified by court actions, we remove these from our ``legitimate'' token set. This also shows that our data set captures projects that occupy the ``middle ground'' with respect to legitimacy, rather than only at extremes of offending and non-offending. The final data set thus consisted of 2,186 tokens (47 of which were the subject of an SEC case in the U.S., i.e., where the individuals/company that created/marketed that token broke securities laws). Our final data set (including our features as described below), can be found here: \url{https://osf.io/xcdz6/?view_only=5a61a06ae9154493b67b24fa4979eddb}.

\subsection*{Features}

Next, we used the Web3 Python package \cite{noauthor_web3py_2023} to extract the bytecode for each of the tokens in our data set. We used an Infura node, which allows users to interface with the Ethereum blockchain through nodes the company runs, using their API\footnote{\footnotesize{https://www.infura.io/}}. Our classification features come from the token smart contract bytecode we collected. We opted to use only code-based features (rather than transaction-based features, for example), following other recent studies which have achieved high levels of performance (including 100\% precision and recall in \cite{RN486}) in classifying Ethereum-based smart contract Ponzi schemes \cite{RN486, RN491, RN492}. Furthermore, using only code-based features allows for classification as soon as smart contracts are deployed \cite{RN395}—rather than waiting to examine the characteristics of associated transactions—and permits analysis of smart contracts with few transactions \cite{RN486}. 
\par
For this initial analysis, our aim was to keep the classifier as simple and as computationally inexpensive as possible. It is also the first classifier for Ethereum-based DeFi securities violations more broadly, so our aim was to obtain a baseline for this novel classification problem in order to determine if it is, in fact, suited to machine learning, rather than improving on the state of the art for previously addressed problems (as \cite{RN491,RN486} and others have done for smart contract Ponzi scheme classification).
\par
EVM bytecode can be computationally ``disassembled'' into its corresponding opcodes. This process is illustrated in Figure \ref{fig4}. Following \cite{RN395}, we included a feature in our classifier for each opcode that appeared in our smart contracts, representing the frequency with which that opcode appeared in any given smart contract. We used the Pyevmasm Python package \cite{RN512} to disassemble each contract’s bytecode into its equivalent opcodes and then used a counter to determine the number of times each opcode appeared in the contract. 

\subsubsection*{Feature Exploration}
Prior to building any classification models, we used elastic net regression (with alpha=0.001, and with the data mean-centered and normalised using scikit-learn's \texttt{StandardScaler})\footnote{\footnotesize{Elastic Net regression combines the ridge penalty (which reduces coefficients of correlated variables) and the lasso penalty (which chooses one of the correlated variables and eliminates the others). The alpha value sets this penalty, with alpha=0 for full ridge regression and alpha=1 for lasso \cite{hastie_introduction_2023}. We chose alpha=0.001 and used the scikit-learn \texttt{StandardScaler} to pre-process our data to enable convergence of our model. The \texttt{StandardScaler} pre-processes the features in a data set by ``removing the mean and scaling to unit variance'' \cite{scikit-learn_developers_sklearnpreprocessingstandardscaler_2023}.}} to explore the importance of our features. Table \ref{tab3} shows the top 10 non-zero coefficients of the model.

\begin{table}[h!]
\caption{Feature exploration using elastic net regression.}
      \begin{tabular}{ll}
      \label{tab3}
        \textbf{Feature} & \textbf{Coefficient} \\ \hline
        SWAP2 & 0.015 \\ \hline
        DUP2 & 0.014 \\ \hline
        SHL & 0.013 \\ \hline
        PUSH30 & 0.012 \\ \hline
        PUSH21 & 0.012 \\ \hline
        PUSH9 & 0.011 \\ \hline
        LOG4 & 0.010 \\ \hline
        MLOAD & 0.008 \\ \hline
        SLT & 0.007 \\ \hline
        RETURN & 0.007\\ \hline
      \end{tabular}
\end{table}

Overall, 55 of the features had non-zero coefficients in our elastic net regression model. However, we note that none of these coefficients were particularly large. This is to be expected because each line of Solidity code ultimately is translated into several opcodes, which means that multiple opcodes could capture the same behavior/action. Table \ref{tab4} describes the opcodes included in Table \ref{tab3}.

\begin{table}[h!]
\caption{Opcode descriptions \cite{RN503}.}
      \begin{tabular}{lp{10cm}}
      \label{tab4}
        \textbf{Opcode} & \textbf{Description} \\ \hline
        SWAP\textit{n} & Exchange first and \textit{n}th stack item\\ \hline
        DUP\textit{n} & Duplicate \textit{n}th stack item\\ \hline
        SHL & ``Shift left''\\ \hline
        PUSH\textit{n} & Put \textit{n}-byte item on the stack \\ \hline
        LOG\textit{n} & Append log record with \textit{n} topics \\ \hline
        MLOAD & Load a word previously saved to memory \\ \hline
        SLT & ``Signed less-than comparison''\\ \hline
        RETURN & Stop code execution and return output data\\ \hline
      \end{tabular}
\end{table}

\subsection*{Classification}
We first used a random forest classifier to attempt to determine if a project was potentially engaging in securities violations. We chose a random forest classifier for the following reasons:

\begin{enumerate}
  \item Research involving data similar to ours achieved the best classification results with a random forest, compared with other classifiers (\cite{RN480}; precision: 96.45\%, recall: 96.79\%, F1-score: 96.62\%).
  \item While initial work on smart contract Ponzi schemes \cite{RN395} has been optimized in later studies (for example, \cite{RN491}), our goal is to achieve a baseline of performance for classifying Ethereum-based DeFi securities violations. Previous work \cite{RN395} found the random forest algorithm performed the best on their data set, when compared with other standard classification algorithms (J48 decision tree and stochastic gradient descent).
  \item Given our primary goal of determining if machine learning methods are suitable for developing a classifier useful for law enforcement investigations, using a model with greater transparency and traceability is most informative.
\end{enumerate}
\par
Given the classification imbalance in our data, we used down-sampling of the majority class to balance it with the minority class. Specifically, we randomly sampled 47 smart contracts from the majority class (i.e., from the n=2,139 legitimate contracts) and ran a random forest classifier on the resulting balanced data set (i.e., 47 violations vs. 47 legitimate tokens). This procedure was repeated 100 times with different random samples, and we report the average performance of these 100 iterations.\footnote{\footnotesize{Under-sampling, combined with properly executed cross-validation, performs well on highly imbalanced data sets \cite{RN529}. While other, related work to ours \cite{RN491} has used the Synthetic Minority Over-Sampling Technique (SMOTE) to train imbalanced data, we chose the more conservative under-sampling method. SMOTE combines majority class under-sampling and minority class over-sampling and synthesizes additional data for the minority class \cite{RN524}.}}
\par
For each iteration, we used 70\% of our data to train our model and 30\% for our test set, following previous work in classifying smart contract Ponzi schemes \cite{RN490}. We calculated accuracy, and weighted precision, recall, and F1-score to evaluate our model \cite{RN523}. We calculated the means of these metrics across our 100 iterations to arrive at our final performance scores. We analyzed the average feature importance across the 100 iterations of our model and then built several subsequent models based on this information.

While, for the aforementioned reasons, we focused on random forest classification, in order to more comprehensively answer our first research question as to machine learning's suitability for this classification task, we needed to build multiple kinds of models, including a simpler approach. For this reason, we also built logistic regression models from our data. As with our random forest classifier, we used down-sampling across 100 iterations, a 70\%-30\% train-test split, and calculated accuracy and weighted precision, recall, and F-1 score metrics. After analyzing feature importance, we constructed further models using different sets of features.\footnote{\footnotesize{We did not build any more complex machine learning models, like neural networks, to answer our first research question because our data set is much smaller than those traditionally used to train deep learning models. Neural networks are much less interpretable than simpler machine learning models \cite{choi_introduction_nodate} and would therefore be less suitable for our purposes (where results could potentially be involved in legal proceedings) in any case.}}
\section*{Results}
\subsection*{Classification}
\subsubsection*{Random Forest}
While we used the results of our elastic net-based feature exploration as input for some of our models, because none of the coefficients were particularly large, we also performed further feature exploration with random forest models. We built our initial classification model using the frequency of all opcodes contained in our data set (a total of 142 features), employing bootstrapped under-sampling to evenly balance the classes in our data set over 100 iterations. As can be seen from the evaluation metrics shown in the top row of Table \ref{tab5}, we achieved satisfactory performance with this model compared to our baseline (50\%). We then calculated the relative importance of the features included in the model, the results of which are reported in Table~\ref{tab6}. 
\begin{table}[h!]
\small
\caption{Random forest model performance with under-sampling.}
      \begin{tabular}{p{6.5cm}p{1cm}ccc}
      \label{tab5}
        & \textbf{Accuracy} &\textbf{Precision} & \textbf{Recall} &\textbf{F1-score} \\ \hline
        RF1: Full-feature model &0.759 &0.757 &0.759 &0.757 \\ \hline
        \textbf{RF2: Top 10 features of model RF1} &\textbf{0.801} &\textbf{0.800} &\textbf{0.800} &\textbf{0.800} \\ \hline
        RF3: Top 3 features of RF1 &0.780 &0.780 &0.780 &0.780 \\ \hline
        RF4: Top feature of RF1 (CALLDATASIZE) &0.777 &0.774 &0.777 &0.774 \\ \hline
        RF5: Non-zero coefficients of EN regression model  &0.731  &0.731 &0.731 &0.731  \\ \hline
        RF6: Top 10 features of RF5  &0.743  &0.742 &0.743 &0.742  \\ \hline
        RF7: Top 3 features of RF5  &0.741  &0.741 &0.741 &0.741 \\ \hline
        RF8: Top feature of RF5 (LT)  &0.747  &0.747 &0.747 &0.747 \\ \hline
        RF9: Top 10 features of EN regression model  &0.690 &0.689 &0.690 &0.689 \\ \hline
     \end{tabular}
\end{table}
\begin{table}[h!]
\tiny
    \caption{Feature importance for random forest models with under-sampling.}
        \begin{tabularx}
        {\columnwidth}{p{1cm} p{.1cm} p{1cm} p{.1cm} p{1cm} p{.1cm} p{.7cm} p{.1cm} p{.7cm} p{.1cm} p{.7cm} p{.1cm} p{.7cm} p{.1cm} }
        \hline
        \multicolumn{2}{c}{\textbf{RF1}} &
        \multicolumn{2}{c}{\textbf{RF2}} &
        \multicolumn{2}{c}{\textbf{RF3}} &
        \multicolumn{2}{c}{\textbf{RF5}} & 
        \multicolumn{2}{c}{\textbf{RF6}} & 
        \multicolumn{2}{c}{\textbf{RF7}} &
        \multicolumn{2}{c}{\textbf{RF9}} 
        \\
        \hline
        \emph{Feature} & \emph{Imp.} & \emph{Feature} & \emph{Imp.} & \emph{Feature} & \emph{Imp.} & \emph{Feature} & \emph{Imp.} & \emph{Feature} & \emph{Imp.} & \emph{Feature} & \emph{Imp.} & \emph{Feature} & \emph{Imp.} \\
        \hline
        CALLDATASIZE & 0.062 & CALLDATASIZE & 0.203 & CALLDATASIZE & 0.340 & LT & 0.077 & LT & 0.165 & LT & 0.357 & SWAP2 & 0.206\\
        \hline
        LT & 0.034 & LT & 0.125 & CALLVALUE & 0.334 & CALLVALUE & 0.075 & CALLVALUE & 0.158 & CALLVALUE & 0.340 & DUP2 & 0.192 \\
        \hline
        CALLVALUE & 0.032 & CALLVALUE & 0.123 & LT & 0.326 & CALLER & 0.046  & CALLER & 0.111 &  CALLER & 0.303 & MLOAD & 0.180\\
        \hline
        SHR & 0.029 & SWAP3 & 0.118 & & & DUP2 & 0.043 & SWAP2 & 0.094 &  &  & RETURN & 0.152 \\
        \hline
        EXP & 0.026 & EXP & 0.110 &  & & SWAP2 & 0.043 & DUP2 & 0.094 &  & & SHL & 0.122 \\
        \hline
        SWAP3 & 0.026 & CALLER & 0.089 & & & MLOAD & 0.039 & MLOAD & 0.088 & & & PUSH21 & 0.061 \\ \hline
        NUMBER & 0.023 & SHR & 0.072 & & & DUP7 & 0.039 & DUP7 & 0.084 & & & SLT & 0.031 \\ \hline
        PUSH5 & 0.021 & NUMBER & 0.063 & & & DUP5 & 0.033 & DUP5 & 0.074 & & & LOG4 & 0.020 \\ \hline
        CALLER	& 0.018	& PUSH5 &	0.055	& & & GT	& 0.033 & GT	& 0.071	& & & PUSH30 & 0.019 \\ \hline
        ADDRESS	& 0.018	& ADDRESS	& 0.041	& & &	DUP11 & 0.032 & DUP11 & 0.061 & & & PUSH9 & 0.018 \\ \hline
        \end{tabularx}
        \label{tab6}
\end{table}
Next, we built models using only the 10 features with the highest importance in our original model, the three most important features, and, finally, the single most important feature (CALLDATASIZE). The weighted precision, recall, and F1-score and accuracy are reported in Table \ref{tab5}. We also built models with all the non-zero coefficients of our elastic net regression model (a total of 55 features), calculated the feature importances for this model, then used this information to build models with the 10 features with the highest importance in this 55-feature model, the top three features, and one with the top feature (LT) of this model. Finally, we built a model using the top 10 non-zero coefficients in our elastic net regression model as our features. We assessed the feature importance for all our further models, and this is reported in Table~\ref{tab6}. Table \ref{tab7} describes the opcodes whose frequency in the smart contracts was determined to be of high importance to the models, which were not previously described.
\begin{table}[h!]
\caption{Additional opcode descriptions \cite{RN503}.}
      \begin{tabular}{lp{10cm}}
        \label{tab7}
        \textbf{Opcode} & \textbf{Description} \\ \hline
        CALLDATASIZE & Retrieve size of ``input data in current environment'' \\ \hline
        LT & ``Less-than comparison'' \\ \hline
        CALLVALUE & Amount deposited by current transaction/instruction \\ \hline
        EXP & ``Exponential operation'' \\ \hline
        CALLER & Get address of caller \\ \hline
        SHR &  “Logical shift right” \\ \hline
        NUMBER & Retrieve block number \\ \hline
        ADDRESS & Retrieve address of account executing transaction \\ \hline 
        GT & ``Greater-than comparison'' \\ \hline
      \end{tabular}    
\end{table}
\par
Using the F1-score as our primary metric, we achieved the best performance with RF2, our 10-feature model built with the top 10 features from our full-feature model (RF1). This model performed relatively well (F-1 score of 80\%) compared to our baseline of 50\%. Three features---the frequencies of CALLDATASIZE, LT, and CALLVALUE---were the most important features across all our random forest models.

\subsubsection*{Logistic Regression}
In order to more fulsomely answer our research question about whether machine learning is appropriate for identifying DeFi projects likely to be engaging in violations of U.S. securities laws, we next built a logistic regression model to see if a simpler model could correctly classify our data. We used the same bootstrapped under-sampling as we did in constructing our random forest models, subsequently calculated feature importance, and built further models accordingly. We report the accuracy, weighted precision, recall, and F-1 scores for these models in Table \ref{tab8} and the feature importance for the top 10 features in Table \ref{tab9}. Table \ref{tab10} describes the opcodes present among the features reported in Table \ref{tab9} which were not previously described.

\begin{table}[h!]
\small
\caption{Logistic regression model performance with under-sampling.}
      \begin{tabular}{p{5.5cm} p{1.2cm} p{1.1cm} p{1.1cm} p{1.3cm}}
        \hline
        & \textbf{Accuracy} &\textbf{Precision} & \textbf{Recall} &\textbf{F1-score}  \\ \hline
        LR1: Full-feature model (using standard scaler) & 0.739	& 0.738 & 0.739	& 0.738 \\ \hline
        LR2: Top 10 features of LR1 & 0.638 & 0.634 & 0.638 & 0.634 \\ \hline
        LR3: Top 8 features of LR1 & 0.639 & 0.634 & 0.639& 0.634 \\ \hline
        LR4: All non-zero coefficients of EN regression (using standard scaler) & 0.725 & 0.724 & 0.725 & 0.724 \\ \hline
        LR5: Top 10 features of LR4 & 0.606 & 0.601 & 0.606	& 0.601 \\ \hline
        LR6: Top 9 features of LR4  & 0.602 &	0.596 &	0.602 &	0.596 \\ \hline
        LR7: Top 10 features of EN regression model  & 0.648	& 0.646 &	0.648 &	0.646 \\ \hline
     \end{tabular}
     \label{tab8}
\end{table}
\begin{table}[h!]
\tiny
    \caption{Feature importance for random forest models with under-sampling.}
        \begin{tabularx}{\columnwidth}{p{.7cm} p{.1cm} p{.7cm} p{.5cm} p{.7cm} p{.1cm} p{.7cm} p{.1cm} p{.7cm} p{.5cm} p{.7cm} p{.1cm} p{.7cm} p{.5cm} }
        \hline
        \multicolumn{2}{c}{\textbf{LR1}} &
        \multicolumn{2}{c}{\textbf{LR2}} &
        \multicolumn{2}{c}{\textbf{LR3}} &
        \multicolumn{2}{c}{\textbf{LR4}} & 
        \multicolumn{2}{c}{\textbf{LR5}} & 
        \multicolumn{2}{c}{\textbf{LR6}} &
        \multicolumn{2}{c}{\textbf{LR7}} 
        \\
        \hline
        \emph{Feature} & \emph{Imp.} & \emph{Feature} & \emph{Imp.} & \emph{Feature} & \emph{Imp.} & \emph{Feature} & \emph{Imp.} & \emph{Feature} & \emph{Imp.} & \emph{Feature} & \emph{Imp.} & \emph{Feature} & \emph{Imp.} \\
        \hline
        SWAP15	& 0.396 & 	PUSH30 & 	0.637 &	CALLVALUE	& 0.043	& SWAP7	& 0.515 & 	BALANCE	& 0.260	& PUSH30 & 	0.701	& PUSH30	& 0.671\\ \hline
        RETURN	& 0.340	& SWAP15 & 	0.514 &	CODESIZE	& 0.484 &	CALLVALUE &	0.514 &	CALLER &	0.023	& SWAP15 &	0.389	& LOG4 &	0.215 \\ \hline
        PUSH30	& 0.299 &	CODESIZE	& 0.443	& EXP	& 0.020 &	PUSH9 &	0.430 &	CALLVALUE	& 0.058	& BALANCE	& 0.215 &	PUSH21	& 0.168 \\ \hline
        CALLVALUE	& 0.287	& PUSH31 &	0.146 &	LOG1	& 0.070 &	RETURN	& 0.427 &	LOG1	& 0.145	& LOG1 &	0.165	& RETURN	& 0.140 \\ \hline
        LOG1 &	0.253 &	LOG1 &	0.102	& PUSH30 &	0.592	& SWAP15 &	0.382	& PUSH30 &	0.706 &	PUSH31	& 0.088 &	SLT	& 0.118 \\ \hline
        PUSH9 &	0.248 &	PUSH9 &	0.091	& PUSH31 &	0.184	& PUSH31	& 0.377 &	PUSH31 &	0.124 &	SWAP7 &	0.064 &	PUSH9	& 0.083 \\ \hline
        PUSH31	& 0.233 &	CALLVALUE	& 0.050 & 	PUSH9	& 0.025	& PUSH30	& 0.328 &	PUSH9 &	0.038 &	CALLVALUE &	0.061 &	DUP2 & 	0.022 \\ \hline
        EXP	& 0.230	& EXP	& 0.022 &	SWAP15	& 0.590 &	LOG1 &	0.271 &	RETURN &	-0.022 &	CALLER &	0.012 &	MLOAD &	0.007 \\ \hline
        SWAP7	& 0.197	& RETURN & -0.030	& & &		CALLER	& 0.266 & 	SWAP15 & 	0.450	& PUSH9	& 0.003 & 	SHL	& -0.031 \\ \hline
        \end{tabularx}
        \label{tab9}
\end{table}

\begin{table}[h!]
\caption{Additional opcode descriptions \cite{RN503}.}
      \begin{tabular}{ll}
        \hline
        \textbf{Opcode} & \textbf{Description} \\ \hline
        CODESIZE & ``Size of code running in current environment'' \\ \hline
        BALANCE & Retrieve account balance \\ \hline
      \end{tabular}
      \label{tab10}
\end{table}
Overall, our logistic regression models performed worse than our random forest models. Our best performing logistic regression models, using the weighted F-1 score as our primary metric, were the models with the most features, namely our 142-feature model (LR1, with an F-1 score of 73.8\%) and our 55-feature model (LR4, with an F-1 score of 72.4\%). Our other logistic regression models performed closer to our baseline performance (50\%).

There is not much overlap in the most important features of our logistic regression and random forest models (besides, of course, in the models built with the top 10 features of our elastic net regression model). EXP was one of the most important features in a handful of both random forest and logistic regression models (RF1, RF2, LR1, LR2, LR3). CALLVALUE was among the top 10 most important features for all our models aside from those built using the top 10 non-zero coefficients of our elastic net regression model. CALLER, which was among the top 10 most important features for five of our random forest models, also had high levels of feature importance in logistic regression models LR4, LR5, LR6. 

Using the weighted F-1 score as our primary metric, we achieved the best performance (an F-1 score of 80\%) with RF2, a 10-feature random forest model. This is, therefore, our final model.

\subsection*{Opcodes}
To better understand the performance of our final model, we compared the frequencies with which the 10 opcodes from our final model occurred in each of our classes (violations and legitimate tokens). We conducted a t-test to assess whether the average frequencies were significantly different.
\begin{table}[h!]
    \caption{Mean comparisons of opcode frequencies and t-test results with Cohen's d effect size.}
    \scriptsize
        \begin{tabularx}{\columnwidth}{p{1.5cm} c p{1cm} c p{1cm} c c p{1cm} l}
        \hline
        & 
        \multicolumn{2}{c}{\textbf{Securities violations}} &
        \multicolumn{2}{c}{\textbf{Legitimate tokens}} &
        \multicolumn{2}{c}{} &
        \multicolumn{2}{c}{}\\
        \hline
        \emph{Opcode} & \emph{Mean} & \emph{St. dev} & \emph{Mean} & \emph{St. dev} & \emph{t-value} & \emph{p-value} & \emph{Effect size \newline (Cohen's d)} & \emph{CI(95\%)} \\
        \hline
        CALLDATASIZE	& 2.745	& 4.245 &	11.387	& 8.192	& -7.210	& $<$0.001	& -1.063 &	[-1.354, -0.772] \\ \hline
        LT	& 9.404	& 9.124 &	18.676	& 14.799 &	-4.277	& $<$0.001	& -0.631 &	[-0.920, -0.341] \\ \hline
        CALLVALUE	& 17.957	& 10.449	& 10.497	& 13.658 &	3.720	& $<$0.001	& 0.549 &	[0.259, 0.838] \\ \hline
        SWAP3 &	24.723	& 17.501 &	37.273 &	25.214	& -3.394 &	$<$0.001 &	-0.500 &	[-0.790, -0.211] \\ \hline
        EXP &	36 &	34.075	& 17.751	& 25.187	& 4.871 &	$<$0.001	& 0.718 &	[0.428, 1.008] \\ \hline
        CALLER	& 16.851	& 9.648 &	11.768	& 11.244 &	3.074	& 0.002 &	0.453 &	[0.164, 0.743] \\ \hline
        SHR	& 0.085	& 0.282 &	0.818 &	1.361 &	-3.688	& $<$0.001	& -0.544	& [-0.833, -0.254] \\ \hline
        NUMBER	& 0.063 &	0.323 &	1.712 &	2.196	& -5.14 &	$<$0.001 &	-0.758 &	[-1.048, -0.468] \\ \hline
        PUSH5	& 0.362	& 1.206 &	2.198 &	2.908 &	-4.321	& $<$0.001	& -0.637 &	[-0.927, -0.348] \\ \hline
        ADDRESS	& 1.255 & 2.982 &	2.529	& 3.255 &	-2.658 &	0.008	& -0.392 &	[-0.681, -0.103] \\ \hline
        \end{tabularx}
        \label{tab11}
\end{table}
The findings from these comparisons are reported in Table \ref{tab11}, and support the analysis of feature importance in our final model. The mean frequencies of each of the features in our final model (reported in Table \ref{tab11}) were significantly different between the securities violations and legitimate token sets, with a much larger effect size for the most important feature (CALLDATASIZE) than for other features. 

\subsection*{Analyzing Solidity Code}

With the aim of better understanding our final model's top three features, we randomly selected five contracts from our set of securities violations and five contracts from our set of legitimate tokens and analyzed their Solidity code. The contracts, the frequencies with which our model's top three features occurred in their code, and the version of Solidity in which their code is written, can be found in Table \ref{tab12}.

\begin{table}[h!]
\caption{Features and Solidity version for contracts analyzed.}
\begin{tabularx}{\columnwidth}{p{3.5cm}lll p{1.5cm}} 
	\hline 
	&
	\multicolumn{3}{c}{\textbf{Frequency of opcodes}} 
        & \textbf{Solidity version} \\ \hline
        & \textbf{CALLDATASIZE} & \textbf{CALLVALUE} & \textbf{LT} & \\ \hline
        \multicolumn{5}{l}{\emph{Violating tokens}} \\ \hline
        Gladius & 5 & 17 & 11 & 0.4.15 \\ \hline
        Tierion Network Token & 1 & 15 & 3 & 0.4.13 \\ \hline
        Dropil & 1 & 16 & 6 & 0.4.18 \\ \hline
        OpportyToken & 1 & 12 & 3 & 0.4.15 \\ \hline
        Boon Tech & 3 & 25 & 18 & 0.4.19 \\ \hline
        \multicolumn{5}{l}{\emph{Legitimate tokens}} \\ \hline
        Sparkle Loyalty & 1 & 19 & 7 & 0.4.25 \\ \hline
        Prometeus & 8 & 12 & 6 & 0.4.23 \\ \hline
        ARC Governance Token & 11 & 1 & 18 & 0.5.0 \\ \hline
        Social Finance & 9 & 22 & 5 & 0.4.23 \\ \hline
        OST & 1 & 26 & 6 & 0.4.17 \\ \hline
\end{tabularx}
\label{tab12}
\end{table}

We used Etherscan,\footnote{\footnotesize{https://etherscan.io/}} the Ethereum blockchain explorer, to obtain the Solidity code for each of these tokens. Next, we used Remix\footnote{\footnotesize{https://remix.ethereum.org/}}---an Ethereum Integrated Development Environment that allows users to write, compile, deploy, and debug Ethereum-based smart contracts, including in virtual environments---to analyze the code. We compiled and deployed each smart contract on the Remix virtual machine. 

We used Remix's ``debugger'' tool to analyze the transactions deploying each section of the compiled contracts. The debugger tool allows a user to examine the opcodes for each transaction chronologically and highlights the corresponding line of Solidity code for each opcode (each line of Solidity code is compiled as several opcodes). It also gives information on the functions that the transaction is interacting with, the local Solidity variables, the Solidity state variables, and other information \cite{remix_debugger_2022, remix_debugging_2022}. However, given our goal of better understanding the features of our final classification model, our analysis focused on the opcode tool.

We examined all elements of the smart contracts involved in their deployment transactions. For each, each time one of our target opcodes appeared, we noted the specific aspect of the token smart contract and the corresponding line of Solidity code.

Though it is difficult to ascertain patterns which may be picked up by our classifier from visual examination of our code, we noticed that four of our five violating contracts (Dropil, Tierion Network Token, OpportyToken, and Boon Tech) had the same line of code which was resolved to the CALLVALUE opcode in the SafeMath portion of the smart contract: \texttt{library SafeMath \{}. In our legitimate token smart contracts, the CALLVALUE opcode was only present in one of the five token contracts (the OST contract) in the SafeMath part of the contract. SafeMath is part of the OpenZeppelin smart contract development library, which allows developers to import standard, vetted, and audited Solidity code for, for example, ERC-20 tokens \cite{openzeppelin_contracts_2023}. The SafeMath library, in particular, provides overflow checks for arithmetic operations in Solidity; arithmetic operations in Solidity ``wrap'' on overflow, which can lead to bugs \cite{openzeppelin_math_2023}. SafeMath solves this issue by reverting transactions that result in operation overflows \cite{openzeppelin_math_2023}.\footnote{\footnotesize{Notably, the SafeMath library was rendered superfluous by Solidity releases 0.8.0 and above (0.8.0 was released in December 2020 \cite{solidity_team_solidity_2020}) \cite{solidity_dev_studio_exploring_2020}. We initially hypothesized that violating tokens could utilize older versions of Solidity than legitimate ones due to the lengthy nature of the U.S. justice process. However, upon further inspection of the Solidity code for each token in our sample, they all rely on Solidity versions between 0.4.13 and 0.5.0 (as shown in Table \ref{tab12}), and all but Gladius include it in their code.}}

Additionally, when subsequently specifying the implementation of SafeMath for various arithmetic operations, the violating tokens use the ``constant'' function modifier, as opposed to the ``pure'' modifier used in the legitimate token smart contracts.\footnote{\footnotesize{In later versions of Solidity, the ``constant'' modifier was changed to ``view'' \cite{the_solidity_authors_contracts_2023}.}} These modifiers dictate whether or not a given function affects the Ethereum global state. The use of ``constant'' indicates that no data from the function is saved or modified, while ``pure'' adds the attribute that the function also does not read blockchain data \cite{nabi_pure_2022}. Essentially, while both attributes specify that the function will not write to the Ethereum state, in the case of ``pure'', the function also does not read state variables \cite{modi_solidity_2018}. The ``pure'' attribute, being stricter about state modification, provides stronger assurance that arithmetic operations resulting in overflow will not (incorrectly) modify the contract's state. It makes sense that the legitimate token contracts would provide this additional security and specificity. 

We previously noted that each line of Solidity code in a smart contract resolves to multiple opcodes. Our smart contract analysis highlights this. In a previous iteration of our model, JUMPDEST was among the top most important features of our model. Various lines of the Solidity code, such as \texttt{Contract BasicToken is ERC20Basic \}} involved both this opcode as well as CALLVALUE. However, we did not notice any distinctions in such lines of code between our violating and legitimate token classes.

CALLVALUE and LT were present numerous times in the aspects of the smart contracts we analyzed with the Remix debugger tool, though, based on our disassembly of these tokens' bytecode, the full frequencies were not present in the portions of the smart contracts we analyzed. The CALLDATASIZE opcode was absent altogether. However, we were unable to execute other transactions in the Remix virtual environment in order to analyze the entirety of the smart contracts (though, we certainly explored a significant portion thereof for each). Furthermore, given our aim in using code-based features to develop a classifier which could be used immediately upon a token contract's deployment, these are the aspects most relevant for our purposes.

\subsection*{Comparing Smart Contracts}
Given our conclusions about the implementation of a particular library in the code being one distinguishing factor between our violating and legitimate token contracts, we sought to delve further into potential code reuse as a reason for our classifier's performance. Prior studies have found that 96\% of Ethereum smart contracts contain duplicative elements (though it is unclear if this is the case in the Ethereum-based DeFi ecosystem specifically)\cite{RN526}. In that sense, if legitimate projects borrow code from other legitimate projects and projects violating securities laws borrow from other violating projects, smart contracts within each class would have a high degree of internal consistency. 
\par

\subsubsection*{Cosine Similarity for Solidity Code}

We used cosine similarities to analyze code reuse among the Solidity code of the smart contracts we analyzed individually. To do this, we tokenized\footnote{``Token'' is used here in the sense of natural language processing, to refer to a portion of text (i.e., a word). It differs from the use of ``token'' in the rest of this article.} the Solidity code using FastText \cite{meta_research_fasttext_2023} and then calculated the cosine similarities between the vectors for each possible combination of the 10 contracts. Cosine similarity measures the level of similarity between two vectors. Cosine similarity is bound to the range -1 to 1; a cosine similarity of -1 means the two vectors are perfectly opposite, 1 means they are identical, and 0 means the two vectors are orthogonal to one another \cite{han_getting_2012}. If code reuse were indeed a possible explanation, we would expect the difference between the cosine similarities within each class and the cosine similarities comparing the violating and legitimate contracts to be more pronounced for violating smart contracts than for legitimate smart contracts. We then compared the means of the cosine similarities each of our token classes. Our results are reported in Table \ref{tab13}.

\begin{table}[h!]
\caption{Cosine similarity of smart contract Solidity code.}
	\begin{tabularx}{\columnwidth}{p{1.8cm} p{1cm} c p{1.1cm} c p{1.5cm} p{2.2cm}} 
	\hline 
	&
	\multicolumn{2}{c}{\textbf{Cosine similarity}} &
	\multicolumn{4}{c}{\textbf{Comparison with inter-class similarity}} \\ \hline
	\emph{Class} & \emph{Mean} & \emph{St. dev} & \emph{t-value} & \emph{p-value} & \emph{Cohen's d} & \emph{CI(95\%)} \\ \hline
	\textbf{Securities violations} & 0.934	& 0.053 &	-0.906 &	0.3711 &	-0.337 &	[-1.067, 0.392] \\ \hline
	\textbf{Legitimate} & 0.962 &	0.0283 &	0.652 &	0.519 &	0.252 &	[-0.506, 1.010]  \\ \hline
         \multicolumn{3}{c}{} & \multicolumn{4}{l}{\textbf{Comparison of violations and legitimate}} \\
         \multicolumn{3}{l}{} & \multicolumn{4}{l}{\textbf{similarities}} \\
         \hline
	\textbf{Inter-class} & 0.951 & 0.048 &	1.361 & 0.191 & 0.625 & [-0.275, 1.523] \\ \hline
	\end{tabularx}
 \label{tab13}
\end{table}

\par

The Solidity code was not significantly different among our classes (at least as per the cosine similarity). We note the high levels of similarity in general among the smart contracts; because of the existence of token standards this is as expected. Based on these results it is unlikely that, in the case of these 10 contracts, code reuse explains our classifier's performance.

\subsubsection*{Cosine similarity of feature-based vectors}
We next compared the opcode frequencies for the smart contracts to one another using cosine similarity to see if the opcodes to which the token Solidity code compiled suggested code reuse could be impacting our classifier's performance. In this experiment, we used the cosine similarity of the vectors of the features rather than the Solidity code itself, as we did in our first cosine similarity experiment, which revealed significant similarities among ERC-20 token smart contracts due to code and token standards. In doing this, we hypothesized that using the opcode frequencies would better capture nuances in the code. We converted the frequencies of the opcodes in each smart contract into a vector and compared them. We calculated the cosine similarity for each possible combination of a) violating smart contracts, b) legitimate smart contracts, and c) violating and legitimate smart contracts (``inter-class''). For each set, we took the average of the calculated cosine similarities, the results of which can be found in Table \ref{tab14}. We also compared the cosine similarities for the set of violating contracts and for the set of legitimate tokens with the inter-class cosine similarities using t-tests; these results are also reported in Table~\ref{tab14}.
\begin{table}[h!]
\caption{Cosine similarity of smart contract opcode frequencies.}
	\begin{tabularx}{\columnwidth}{p{1.8cm} p{1cm} c p{1.1cm} c p{1.5cm} p{2.2cm}} 
	\hline 
	&
	\multicolumn{2}{c}{\textbf{Cosine similarity}} &
	\multicolumn{4}{c}{\textbf{Comparison with inter-class similarity}} \\ \hline
	\emph{Class} & \emph{Mean} & \emph{St. dev} & \emph{t-value} & \emph{p-value} & \emph{Cohen's d} & \emph{CI(95\%)} \\ \hline
	\textbf{Securities violations} & 0.040 &	0.024	& -8.561	& $<$0.001	& -0.262 &	[-0.322, -0.202] \\ \hline
	\textbf{Legitimate} & 0.054	 & 0.062 &	7.719 &	$<$0.001 & 0.025 & [0.019, 0.031]  \\ \hline
         \multicolumn{3}{c}{} & \multicolumn{4}{l}{\textbf{Comparison of violations and legitimate}} \\
         \multicolumn{3}{l}{} & \multicolumn{4}{l}{\textbf{similarities}} \\
         \hline
	\textbf{Inter-class} & 0.052 &	0.048 &	7.381 &	$<$0.001	& 0.225 &	[0.165, 0.284]\\ \hline
	\end{tabularx}
 \label{tab14}
\end{table}
\par
Our results show that the legitimate smart contracts' opcodes are slightly more similar to each other than the violating contracts are (0.054 and 0.040, respectively). The violating contracts' opcodes are less similar to each other than they are to the legitimate contracts' (at least per the cosine similarities, which are 0.052 for inter-class cosine similarity and 0.040 for the violating class). This suggests that there may be slightly more code reuse among legitimate contracts than violating ones. However, the cosine similarity for both groups is rather small, as are the effect sizes, which suggests that code reuse overall is unlikely to be the primary reason for our classifier's performance (though this does not preclude the possibility of certain elements of the code---such as the use of the SafeMath library---being at least partially responsible). 

\section*{Discussion}
This article sought to determine if it is useful to build a machine learning classifier to detect DeFi projects engaging in various types of securities violations from their tokens’ smart contract code. Governments are currently struggling with how to manage fraud in the DeFi ecosystem, particularly because these platforms do not require KYC; a classifier could serve as a triage measure. In addition, we make available a new data set with a verified ground truth. Our research is also novel in its use of ERC-20 token smart contract code to attempt to detect fraud across the Ethereum-based DeFi ecosystem and deeper analysis of the features of our final model. Finally, our research contributes to the existing body of work on financial markets, as well as its practical applications, in that predicting and detecting fraud is crucial for financial stability, market integrity, and market efficiency, particularly in the cryptocurrency space \cite{shams_detection_2021}.
\par
Ultimately, we found that DeFi securities violations are a detectable problem. We developed a suitable starting point for this classification problem that performed much better than the baseline (80\% F-1 score against a baseline of 50\%). Our performance was not as high as others' models for similar classification problems, however, as we describe below, some of this may be due to over-fitting and the data sets used in other work. Previously developed baseline models for related classification problems exhibited performance more in line with ours.
\par
With regard to our second research question, further analysis at the feature level indicated that the success of our model may, in part, be related to the state-based attributes of functions in the SafeMath library.

\subsection*{Comparisons with prior research}
Since ours is the first study to attempt to classify DeFi securities violations more broadly, we are unable to compare our model’s performance with previous studies. We do note that other studies which successfully built high performance classifiers for a related classification problem used more complex methods, which improved upon several previous studies addressing the same classification problem  \cite{RN486}. In addition, many previous studies, like ours, also used data with quite imbalanced classes, but they did not always account for this in building their models. Fan et al. \cite{RN491}, in particular, criticize previous work, including \cite{RN395}, on this basis. This may make the high performance metrics reported by some other studies slightly misleading. For example, only 3.6\% of the smart contracts in the Chen et al. \cite{RN1541} data set are Ponzi schemes. 
\par
In contrast to previous studies, ours considers whether machine learning classification is necessary or whether a simpler model may suffice to solve the same problem. Previous studies have found that models built with, for example, logistic regression, have been much less effective than more complex ones \cite{RN480}. We also found this to be the case.

Chen et al. \cite{RN486} note the overall lack of interpretability of the results of classifiers with code-based features. The opcodes themselves have no obvious interpretation with respect to illegal activity, but, equally, there are not any opcodes which would offer a straightforward interpretation—i.e., there is no “STEAL" opcode or similar. The opcodes whose frequencies were among the most important features in our model were CALLDATASIZE, LT, CALLVALUE, SWAP3, EXP, CALLER, SHR, NUMBER, PUSH5, and ADDRESS. The only way to draw definitive conclusions from opcode-based features is to dissect the Solidity code from which they were compiled. Though it would have been impractical to dissect all 2,186 of the smart contracts in our data set, we gleaned some insight about our three most important features from a selection thereof, namely that developers of violating tokens may implement the SafeMath library differently in their code. In particular, they appear to use the ``constant'' modifier when describing how arithmetic operation overflows should be handled, which offers weaker assurances about the lack of state modification by these functions. It is, therefore, intuitive, for these reasons, that the use of the ``pure'' function would be associated with the legitimate tokens in this case. However, we note that, our analysis of transactions deploying the compiled Solidity code for certain contracts does not capture all of the opcodes whose frequencies were among the top 10 most important features in our model.  Future research could also utilize other frameworks for analyzing token transfer behavior from token bytecode, such as TokenAware \cite{10.1145/3560263}. This would be particularly useful as well in instances where the contract Solidity code is not publicly available. 

\subsection*{The use of code-based features}

Like previous work \cite{RN395, RN486}, we emphasize the usefulness of our classifier immediately upon deployment of the smart contract to the Ethereum blockchain and regardless of how many wallets interact with it. This is one of the key advantages of using code-only features for classification rather than transaction- or account-based features. This makes such a model not only useful as a retroactive tool for investigators, but also to prevent future fraud. It also enables investigators to monitor projects which may engage in securities violations in the future. Since investigations and prosecutions take such a long time (upwards of several years for complex cases), it is important for prosecutors to be able to begin gathering evidence as early as possible. However, we acknowledge that code-based analysis is merely one technique among many; future research could explore alternatives.

\subsection*{Potential applications of our model}

Our results point to the value of exploring the use of computational triage systems in the enforcement process. This is particularly important given that U.S. enforcement agencies seem to be relying heavily on submissions to their whistleblower programs \cite{RN528}; a computational model could reduce reliance on whistleblowers and also avoid the government needing to pay out a portion of funds successfully recovered to whistleblowers (which can be millions of dollars \cite{RN527}). 
\par
A classifier of the type we examine here would be more useful as a triage measure rather than a source of evidence due to issues surrounding the admissibility of machine learning-generated evidence in U.S. courts and also the risk for mis-classification. There may be questions about its admissibility under the Fifth Amendment, the Sixth Amendment, and the Federal Rules of Evidence, however legal scholars do not, ultimately, consider them impediments to its admission \cite{RN497}.\footnote{\footnotesize{Though a complete discussion of the admissibility of machine learning evidence is outside of the scope of this paper, we provide a brief introduction here. The Fifth Amendment relates to an individual's right to due process (this could arise in the context of the ``black box'' of machine learning calculations) and the Sixth Amendment includes the Confrontation Clause. This ``black box'' not only refers to inexplicable machine learning algorithms but also lay people’s likely lack of understanding about how these algorithms work. The Confrontation Clause requires experts to testify in person and submit to cross-examination. However, this is unlikely to be an issue, as the testimony of a machine learning expert should be satisfactory. The Federal Rules of Evidence around relevance, prejudice, and authenticity may be pertinent as well. Lawyers must further prove the accuracy of the evidence \cite{RN1543}. Some argue that under Rule 702 and \textit{Daubert v. Merrell Dow Pharmaceuticals}, machine learning evidence would be admissible as expert testimony. Through \textit{Daubert}, the court developed a set of four considerations for evaluating expert testimony \cite{RN497}.}} Even if it is admissible, however, questions about its weight in court remain. In particular, explaining such evidence to a judge and jury (especially the ``black box'' calculations involved in developing machine learning models) may lead it being discounted. There is variation in levels of trust among jurors in machines in general and jurors must further trust the expert testimony which explains the machine learning tool \cite{RN497}. This is exacerbated by the need for prosecutors to already explain complex concepts related to cryptocurrency in these cases. A machine learning-based tool would likely lead investigators to more compelling transaction-based evidence or qualitative evidence (for example, marketing material) which can be more easily understood by a judge and jury and which has been effective in prosecuting cryptocurrency-based financial criminal offenses in the past (See \cite{RN520, RN519}).
\par
Considering the ambiguity around the Hinman standard\footnote{\footnotesize{The ``Hinman standard'' refers to William Hinman's 2018 speech which considered the level of decentralization of a project critical to determining whether it should be classed as a security \cite{RN507}.}} in determining whether a project is sufficiently decentralized to avoid being classed as a security, a machine learning model could also serve as an additional tool in developers’ arsenal for determination thereof. Finally, we also perceive a machine learning model as potentially useful for people who are interested in participating in the DeFi ecosystem, as a way for them to research the validity of new projects to help protect themselves from fraud.

\subsection*{Limitations and future research}

Our research may suffer from various limitations. The first is the potential for overfitting our model, particularly in the face of imbalanced data \cite{RN491}. However, since we do not have a separate data set with known labels on which to test our model, overfitting could remain an issue despite our mitigation efforts. Ultimately, we chose a verified ground truth data set that was much smaller than our set of legitimate tokens. We do note that our data set may be making this classification problem simpler than it is in reality. We chose a list of generally reputable projects for our legitimate token set and those which are subject to government enforcement action for our securities violations. In this sense, we believe that, given the experimental nature of DeFi (and the high risk appetite of its participants), and the initial inclusion of a few violating tokens in the legitimate set, this set captures projects that exist in the middle, rather than at only extremes of offending and non-offending, which may be harder to classify. However, it would still be useful for future research to develop more data sets of DeFi securities violations to further test and refine our models, as necessary, using more advanced sequential machine learning techniques. As the number of verified violations increases, future research could also explore DeFi securities violations using more granular classes. These classes would show the different security patterns of the various types of securities violations and could further aid in prevention and detection efforts. It would also be useful for future research to compare the use of code-based features with models using account-based features or a combination thereof.  

We note that, given the seeming importance of the implementation of the SafeMath library, in part, to our classifier's performance, it may be the case that this classifier is less effective in classifying tokens created with Solidity versions 0.8.0 or later. However, this code did not account for all of our most important features. Future research could examine this, again, using an expanded data set. 

We accept that there are some limitations around the use of a classifier like this in practice. Our future research will specifically explore how we could use a classifier like this to conduct an investigation and build a viable legal case. This will involve performing manual, in-depth analysis on tokens flagged when applying our classifier to other data sets and interactions with their smart contracts (similarly to Xia et al.'s \cite{RN480} work on Uniswap scam tokens). 

Chen et al. \cite{RN486} raise the issue of bad actors using adversarial obfuscation methods to trick classifiers like the one we propose. We did not explicitly account for this possibility in building or model, nor did we test our model against known obfuscation techniques. However, this may be a useful avenue for future research to explore.

Further analysis of violating contracts, for example, using methods for analyzing token operational behavior from bytecode (such as TokenAware, which has successfully been applied to discrete instances of fraud \cite{10.1145/3560263}) would be fruitful. We encourage other scholars to pursue this line of research using our publicly available data set.

Finally, the jurisdictional focus of our research was limited in scope due to the nature of our data set. Legislation relating to cryptocurrencies is frequently changing; in particular, the European Union recently passed the Regulation on Markets in Cryptoassets (MiCA)\cite{scicluna_mica_2023}\footnote{\footnotesize{Currently, the only implemented EU regulations that apply to cryptocurrencies relate to money laundering. The EU's securities and investment regulations do not currently apply \cite{kolinska_cryptocurrencies_2022}.}}. More generally, a global approach, incorporating legislation from multiple jurisdictions, would be a potential future goal. This would, of course, bring challenges, however, since the legality of particular applications—equivalent to the labels of training data, in technical terms—may vary across jurisdictions and change over time. Overcoming this challenge may require the adaptation of approaches from other domains.

One further prospect is that the output of a classification system such as this may be useful in detecting flaws or risks in novel DeFi functions. While the model would be trained on violations that had previously been detected, it is possible that cases may be identified as being risky even if they do not correspond to known flaws, but rather because they have underlying similarities to existing cases. In such a situation, manual inspection of cases identified as potentially fraudulent may offer insight into new forms of offending. This type of work would usefully contribute to the goal of preventing DeFi frauds and associated losses to individuals. Research on the prevention of such offenses is a crucial complement to the detection work presented in this paper.

\section*{Conclusions}
Our final model achieves good performance (80\% F-1 score against a baseline of 50\%) in classifying DeFi-based securities violations based on ten features from the projects’ tokens’ smart contract code: the frequencies with which the CALLDATASIZE, LT, CALLVALUE, SWAP3, EXP, CALLER, SHR, NUMBER, PUSH5, and ADDRESS opcodes occurred in the contract. We achieved higher performance with this random forest model than with logistic regression models, leading us to conclude that this classification problem is, indeed, well-suited to machine learning. Our research is novel in its deeper analysis of the opcode-based features responsible for the performance of our classifier. Though this does not account for all features, the implementation of the SafeMath library in the token smart contracts appears to play a role. Despite the seeming influence of this aspect of the token smart contract code on our classifier, cosine similarity analyses did not suggest code reuse overall was the primary reason for its performance. Overall, a computational model like ours would be highly useful for investigators as a triage tool but could be circumvented by nefarious developers in the future. It is important, therefore, to augment any model as further DeFi projects engaging in securities violations are revealed. This work constitutes the first classifier of securities violations overall in the emerging and fast-growing Ethereum-based DeFi ecosystem and is a useful first step in tackling the documented problem of DeFi fraud. Our work also contributes a novel data set of DeFi securities violations with a verified ground truth and connects the use of such a classifier with the wider legal context, including how law enforcement can use it from the investigative to prosecution stages of a case.


\begin{backmatter}

\section*{Declarations}

\section*{Availability of data and materials}
The datasets generated and analysed during the current study are available in the OSF repository, \href{https://osf.io/xcdz6/?view\_only=5a61a06ae9154493b67b24fa4979eddb}{Detecting Defi Securities Violations}. 

\section*{Competing interests}
The authors declare that they have no competing interests.

\section*{Funding}
This work was funded by the [redacted for blind review]. 

\section*{Author's contributions}
AT: conceptualization, data collection, analysis, interpretation, and drafting the final manuscript. BT, TD: conceptualization, study design, and feedback on the manuscript. All authors have reviewed the final manuscript.

\section*{Acknowledgements}
The authors also thank Antonis Papasavva and Antoine Vendeville for their contributions to our code.
  
\section*{List of abbreviations}
\begin{itemize}
	\item \textbf {DeFi:} Decentralized finance
	\item \textbf{dApps:} Decentralized applications
	\item \textbf{Opcode:} Operational code
	\item \textbf{EVM:} Ethereum Virtual Machine
	\item \textbf{Dexes:} Decentralized exchanges
	\item \textbf{KYC:} Know Your Client
	\item \textbf{SEC:} Securities and Exchange Commission
	\item \textbf{FINRA:} Financial Industry Regulatory Authority
	\item \textbf{ICO:} Initial Coin Offering
	\item \textbf{ERC-20:} Ethereum Request for Comment
	\item \textbf{BA:} Blockchain Association
	\item \textbf{SMOTE:} Synthetic Minority Over-Sampling Technique
\end{itemize}

\bibliographystyle{bmc-mathphys} 
\bibliography{references}      







\end{backmatter}

\end{doublespace}
\end{document}